\documentclass[10pt,twocolumn,letterpaper]{article}
\usepackage{authblk}
\usepackage{cvpr}      

\usepackage{amsmath,bm}
\usepackage{multirow}
\usepackage{tcolorbox}
\usepackage{graphicx}
\tcbuselibrary{breakable}
\usepackage{float}
\usepackage[dvipsnames]{xcolor}
\usepackage{booktabs} 
\usepackage{array}
\usepackage{makecell}
\usepackage{ragged2e} 
\usepackage{footmisc}
\definecolor{cvprblue}{rgb}{0.21,0.49,0.74}
\usepackage[pagebackref,breaklinks,colorlinks,allcolors=cvprblue]{hyperref}

\def\paperID{*****} 
\def\confName{CVPR}

\title{SpatialCoT: Advancing Spatial Reasoning through Coordinate Alignment and Chain-of-Thought for Embodied Task Planning}

\author{Yuecheng Liu\textsuperscript{*}}
\author{Dafeng Chi\textsuperscript{*}}
\author{Shiguang Wu\textsuperscript{*}}
\author{Zhanguang Zhang}
\author{Yaochen Hu}
\author{Lingfeng Zhang}
\author{Yingxue Zhang}
\author{Shuang Wu}
\author{Tongtong Cao}
\author{Guowei Huang}
\author{Helong Huang}
\author{Guangjian Tian}
\author{Weichao Qiu}
\author{Xingyue Quan}
\author{Jianye Hao}
\author{Yuzheng Zhuang\textsuperscript{†}}
\affil{Huawei Noah's Ark Lab}

\begin{document}
\maketitle

\let\thefootnote\relax\footnotetext{\begin{raggedright}*: Equal contribution. Email:  \{liuyuecheng1, chidafeng1, wushiguang\}@huawei.com\end{raggedright}}
\let\thefootnote\relax\footnotetext{†: Corresponding author. Email: zhuangyuzheng@huawei.com}

\begin{abstract}
Spatial reasoning is an essential problem in embodied AI research. Efforts to enhance spatial reasoning abilities through supplementary spatial data and fine-tuning have proven limited and ineffective when addressing complex embodied tasks, largely due to their dependence on language-based outputs. While some approaches have introduced a point-based action space to mitigate this issue, they fall short in managing more intricate tasks within complex environments. This deficiency arises from their failure to fully exploit the inherent thinking and reasoning capabilities that are fundamental strengths of Vision-Language Models (VLMs). To address these limitations, we propose a novel approach named \textbf{SpatialCoT}, specifically designed to bolster the spatial reasoning capabilities of VLMs. Our approach comprises two stages: \textit{spatial coordinate bi-directional alignment}, which aligns vision-language inputs with spatial coordinates, and \textit{chain-of-thought spatial grounding}, which harnesses the reasoning capabilities of language models for advanced spatial reasoning. We evaluate SpatialCoT on challenging navigation and manipulation tasks, both in simulation and real-world settings. Experimental results demonstrate that our method significantly outperforms previous state-of-the-art approaches in both tasks. Project page: \href{https://spatialcot.github.io}{https://spatialcot.github.io}.
\end{abstract}

\section{Introduction}
\begin{figure}
    \centering
    \includegraphics[width=1.0\linewidth]{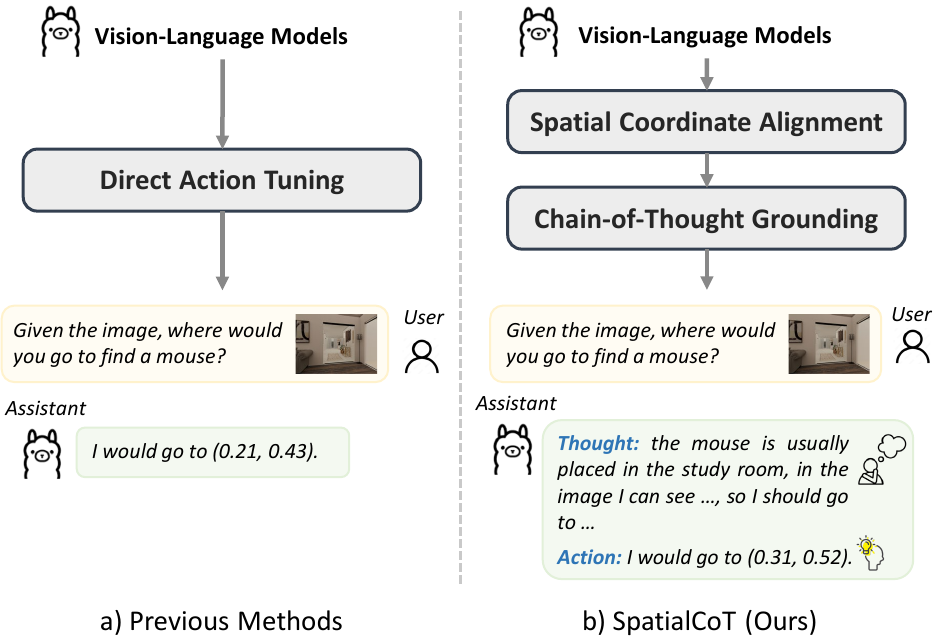}
    \caption{\textbf{Comparison between \textbf{SpatialCoT} and previous methods}. a) Previous methods usually directly output the action based on the language instruction. b) \textbf{SpatialCoT} enhances action generation quality by effectively leveraging the reasoning capabilities of VLMs. This is achieved through a two-stage finetuning process involving spatial coordinate alignment and chain-of-thought spatial grounding.}
    \label{fig:brief-framework}
\end{figure}

Spatial reasoning, defined as the cognitive ability to visualize, manipulate, and comprehend the relationships between objects, is fundamental for performing everyday tasks such as navigating environments, assembling furniture, and organizing items on a table. Recent advancements in Large Language Models (LLMs)~\cite{llm-overview} and Vision Language Models (VLMs)~\cite{vlm-overview} have established spatial reasoning as a crucial tool for embodied AI researchers in completing embodied tasks. However, most VLMs~\cite{PaLI, BLIP-2, Palm-e, minigpt4, nextgpt} are trained on standard 2D images and text datasets, which lacks the information necessary for understanding spatial relationships, thereby limiting their spatial reasoning abilities. Some works have attempted to  enhance these capabilities by integrating additional spatial data and refining the models accordingly \cite{spatialVLM, spatialRGPT, embosr}. Nevertheless, these efforts primarily focus on language-based reasoning, resulting in models that produce only coarse-grained reasoning results. This constraint significantly limits the range of embodied applications, particularly within tasks that demand sophisticated action decisions. For example, when a robot receives an instruction to set up a table and the model generates subtasks such as \emph{``1) Put the cup on the top left to the right of the plate., 2) Put ...''}, these commands are easy for humans to interpret. However, for a low-level policy, often implemented using a smaller model such as decision transformer~\cite{decision-transformer} or diffusion model~\cite{diffusion-policy, Scaling-Up-and-Distilling-Down}, determining the correct placement of the cup while avoiding collisions with other objects presents a significant challenge, rendering the command ambiguous.

Recent work, such as RoboPoint~\cite{robopoint} and RoboSpatial~\cite{robospatial}, addresses this issue by introducing a point-based action space. For instance, given a spatially related instruction like  ``\textit{left of bowl and on the tarp}'', the model generates one or several points on the input image to indicate the location or region described. While RoboPoint performs satisfactorily on several basic spatial reasoning tasks, such as object reference and free space reference, it exhibits notable limitations. 
The model's approach of directly translating language instructions into points bypasses the inherent language-based reasoning capabilities of VLMs. As a result, this method overlooks the core strengths of large language models, thereby constraining the model's ability to manage complex tasks. Particularly, it struggles with those tasks requiring detailed or multi-step reasoning in intricate environments.

Simultaneously, chain-of-thought (CoT) prompting \cite{CoT} and its extensions have emerged as a prevalent methodology for researchers to tackle complex tasks using large language models~\cite{tree-of-thought, code-as-policies, react} or vison-language models~\cite{socratic-models, multimodal-CoT} . This approach is also being explored within the domain of embodied AI~\cite{inner-monologue, ECoT-openVLA}. In these studies, models are instructed to articulate their thought processes prior to arriving at a final answer. While these works focus on language-based planning, the challenge of leveraging these thought processes to generate fine-grained actions remains largely unaddressed.

To overcome the limitations of existing methods, in this work, we propose a novel approach, termed \textbf{SpatialCoT}, to enhance the spatial reasoning capabilities of vision-language models for embodied task planning (shown in Figure~\ref{fig:brief-framework}-b). The approach comprises two stages: 1) \textbf{\emph{spatial coordinate bi-directional alignment}}: This stage involves the explicit alignment between vision-language inputs and coordinates, thereby enabling the model to better comprehend and generate coordinate-based responses. We introduce a bi-directional alignment mechanism to further reinforce this process. 2) \textbf{\emph{chain-of-thought spatial grounding}}: In this phase, we enhance spatial reasoning by explicitly utilizing the language-based reasoning capabilities of vision-language models, rather than directly generating coordinate-based actions in an end-to-end manner. This approach significantly improves the model's ability to handle more complex tasks in intricate environments. Additionally, we introduce a  pipeline to automatically generate data with high-quality rationales for model fine-tuning, which substantially reduces data acquisition costs.

Differing from prior studies in spatial reasoning research, which typically conduct open-loop evaluations on offline visual question answering (VQA) datasets, this paper adopts a more challenging setting by performing closed-loop evaluations within  simulators and real world. The evaluation tasks encompass both navigation and manipulation, each presenting greater challenges compared to those in earlier works. The experimental results demonstrate that our model significantly outperforms previous state-of-the-art methods in both tasks.

In summary, the key contributions of this paper are:
\begin{itemize}
    \item A novel approach, \textbf{SpatialCoT}, designed to enhance the spatial reasoning abilities of vision language models for fine-grained action generation, comprising two stages: \emph{spatial coordinate bi-directional alignment} and \emph{chain-of-thought spatial grounding}. The approach explicitly leverages the inherent language-based reasoning capabilities of vision language models, significantly improving performance on complex embodied tasks.
    \item A pipeline that enables the automatic collection of data with high-quality rationale, substantially reducing data acquisition costs for model fine-tuning. 
    \item State-of-the-art results on challenging embodied planning tasks, including both navigation and manipulation.
\end{itemize}

\section{Related Work}

\begin{figure*}
    \centering
    \includegraphics[width=\textwidth]{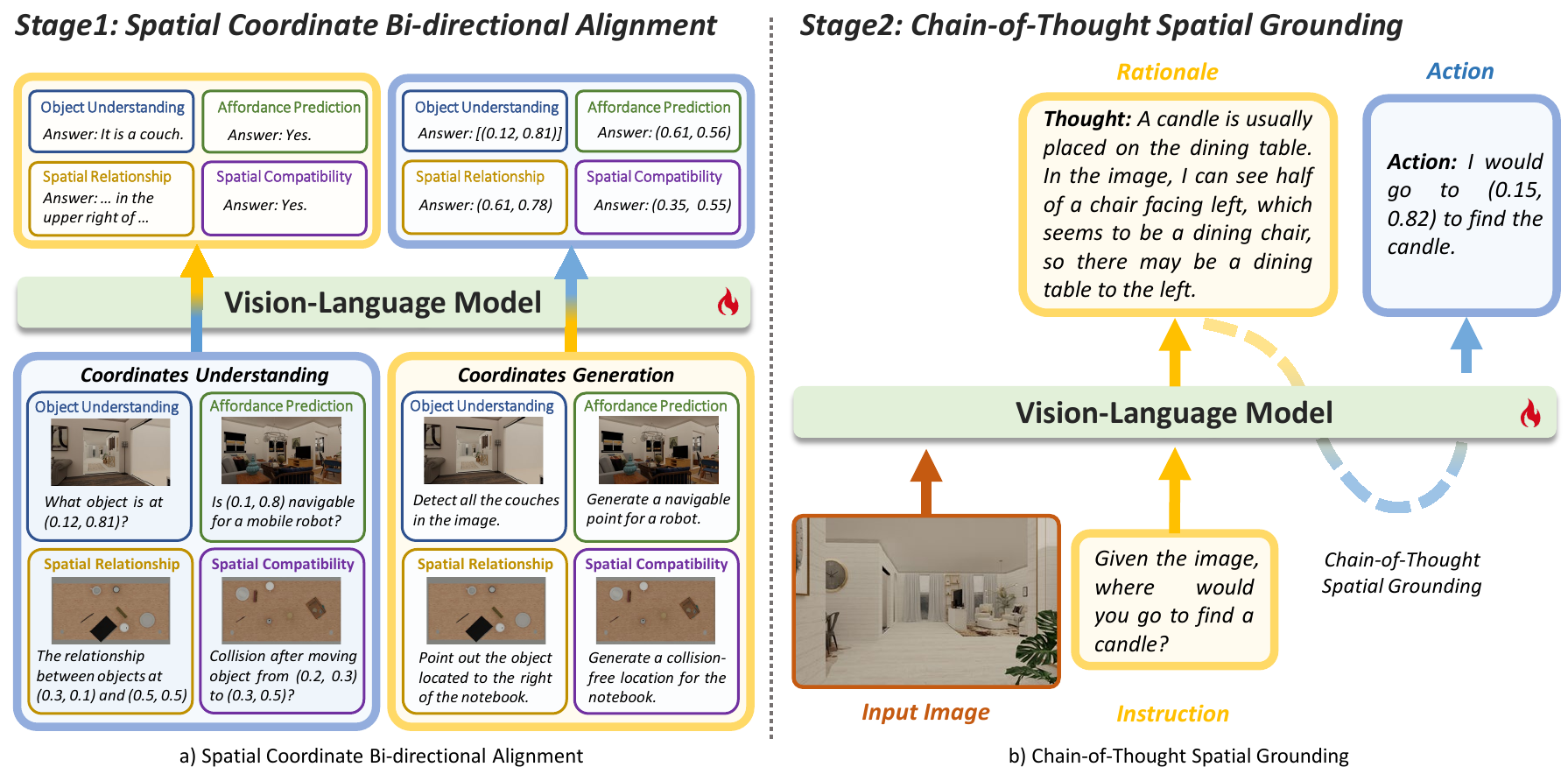}
     \caption{\textbf{Overview of SpatialCoT, comprising two core stages.} a) \textbf{Spatial coordinate bi-directional alignment}, which involves translating coordinates to language (indicated by the blue to yellow arrow on the left) and language to coordinates (indicated by the yellow to blue arrow on the right). b) \textbf{Chain-of-thought spatial grounding}: the model first performs comprehensive thinking by generating a language-based rationale, and then grounds it in coordinate-based actions (yellow to blue dashed line), significantly improving the model's performance in complex spatial reasoning tasks.}

    \label{fig:detailed-framework}
\end{figure*}

\subsection{Spatial Reasoning}

Spatial reasoning is a crucial capability for vision language models and is included in numerous VQA benchmarks \cite{GQA, clevrer, GSR-bench, thinking-in-space}. However, most VLMs~\cite{PaLI, BLIP-2, Palm-e, minigpt4, nextgpt} are predominantly trained on 2D images paired with text, which lack sufficient spatial data. As a result, their spatial reasoning abilities are limited. To address this issue, some works, such as SpatialVLM~\cite{spatialVLM} and SpatialRGPT~\cite{spatialRGPT} have been developed to enhance the spatial reasoning of VLMs by collecting spatially-related question-answering data and fine-tuning the models on them. Recent works has further extended spatial reasoning to generate more fine-grained actions by introducing a point-based action space \cite{robopoint, robospatial}. Given a spatially related instruction, RoboPoint \cite{robopoint} outputs one or several points located on the input image to indicate the location or region described in the instruction, a process the authors call “\emph{spatial affordance prediction}.” Following the idea of RoboPoint, RoboSpatial \cite{robospatial} makes further improvements by introducing more types of data. However, these studies primarily focus on establishing a direct mapping between language instructions and corresponding points, thereby neglecting the incorporation of language-based reasoning capabilities of VLMs. This limitation hampers the model's proficiency in managing more challenging tasks, particularly those that necessitate intricate or multi-step reasoning.

\subsection{Embodied Chain-of-Thought}
Chain-of-thought \cite{CoT} and its extensions have become key techniques in large language models \cite{tree-of-thought, react, GoT} and vision-language models \cite{socratic-models, multimodal-CoT} enhance problem-solving abilities by guiding them through a series of logical steps. Instead of directly providing an answer, CoT prompts the model to break down a problem into smaller, manageable parts, enabling more systematic and accurate reasoning. This technique has been explored in previous works to tackle complex embodied tasks~\cite{inner-monologue, code-as-policies}. For example, Inner-Monologue \cite{inner-monologue} prompts the model to leverage feedback from the environment to create an “inner monologue” that helps LLMs process and plan more effectively. CaP \cite{code-as-policies} models the thinking process of the model into the form of code generation. However, these approaches primarily focus on language-based (i.e., coarse-grained) planning, and we argue that this thinking process can also be beneficial for fine-grained spatial reasoning.

\section{Our Method}

    Our method consists of two fundamental stages, as illustrated in Figure~\ref{fig:detailed-framework}. The first stage, termed \emph{spatial coordinate bi-directional alignment}, equips the vision-language models with the capability to understand and generate coordinates. The second stage, \emph{chain-of-thought spatial grounding}, enables the model to engage in comprehensive reasoning and to translate this reasoning into coordinate-based actions, leveraging the alignment ability developed in the first stage. The following sections will provide a detailed explanation of each stage.

\begin{figure*}
    \centering
    \includegraphics[width=\textwidth]{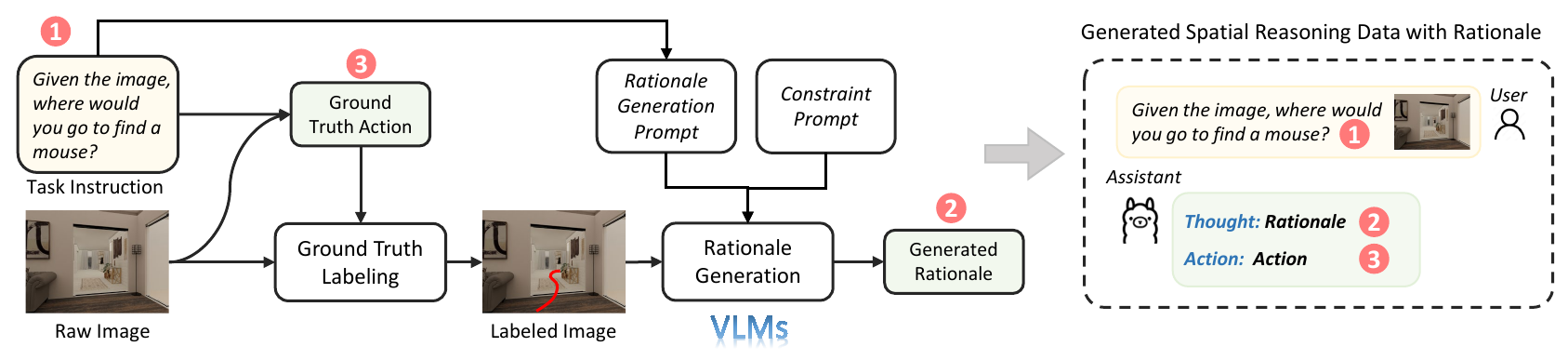}
    \caption{Data collection pipeline for chain-of-thought spatial grounding}
    \label{fig:rationale-collection}
\end{figure*}

\subsection{Spatial Coordinate Bi-directional Alignment}
\label{section:spatial-coordinate-alignment}

Previous studies~\cite{robopoint, robospatial} have attempted to leverage additional VQA data, such as object references, as co-training data. However, the organization of this data in these works is often lacking, and its potential to enhance the model’s spatial reasoning capabilities remains underutilized. In this work, we propose an explicit alignment of vision-language data with coordinates, which will significantly aid the model in understanding and generating coordinate-based inputs and outputs. Unlike previous studies, we introduce a \emph{bi-directional} alignment framework to strengthen this process. Let $\textbf{X}_v$ represent an image, $\textbf{X}_\text{lang}$ represent language-only text (without coordinates), $\textbf{X}_\text{coor}$ represent text containing one or more coordinates, and $f_{\boldsymbol{\theta}}(\cdot)$ represent a auto-regressive VLMs parameterized by $\bm{\theta}$. For each type of data, we design two different forms. The first form takes an image and a text-based instruction with coordinates in it, and the model should output the corresponding information about the given coordinates described in the instruction (equation \ref{equation-1}). For example, \textit{“Question:\textless image\textgreater  What is the object located at (0.81, 0.90)? Answer: chair”}. For the second form, the model is given an image together with language-only instructions (without coordinates), and the model is asked to generate one or several coordinates to point out the location or region described in the instruction ( equation \ref{equation-2}). For example, \textit{“Question: \textless image \textgreater 
 Give the locations of all the chairs in the image. Answer: [(0.12, 0.31), (0.31, 0.35), …]”}.
\begin{align}
    \label{equation-1} 
   & [\textbf{X}_v, \textbf{X}_\text{coor}] \xrightarrow[]{f_{\boldsymbol{\theta}}(\cdot)} \textbf{X}_\text{lang}\quad\text{coordinates understanding} \\
   \label{equation-2}
    & [\textbf{X}_v, \textbf{X}_\text{lang}] \xrightarrow[]{f_{\boldsymbol{\theta}}(\cdot)} \textbf{X}_\text{coor}\quad\text{coordinates generation} 
\end{align}

To achieve a more comprehensive alignment between vision-language and coordinates, we introduce various types of data, which can be categorized into four distinct groups: 
\begin{itemize}
    \item \textbf{Object Understanding}: This involves matching natural language descriptions with specific visual content in images. This process is also referred to as visual grounding \cite{kosmos-2}. Essentially, it aims to identify and locate objects within an image based on a given textual description.
    \item \textbf{Affordance Prediction} Affordance prediction refers to identifying and predicting the possible actions that an object or environment allows. For example, determining which areas are navigable for a mobile robot without collision with obstacles, or understanding how to grasp or operate certain objects.
    \item \textbf{Spatial Relationship}: This type of data pertains to understanding the relationships between objects based on the layout of the environment.
    \item \textbf{Spatial Compatibility}: This type of data aims to enhance models' abilities to understand and predict the compatibility between objects.
\end{itemize}

We illustrate the spatial coordinate bi-directional alignment stage in Figure~\ref{fig:detailed-framework}-a). Detailed examples of the data, including prompts and responses, can be found in Appendix~\ref{appendix:dataset-examples}.

\subsection{Chain-of-Thought Spatial Grounding}
Unlike previous works~\cite{robopoint, robospatial} that directly output coordinate-based actions given the language instruction (equation~\ref{equation-3}), this work aims to explicitly utilize the language-based reasoning abilities of VLMs to address complex spatial reasoning tasks. Drawing inspiration from ReAct \cite{react}, we generate additional data where the output is divided into two components (equation~\ref{equation-4}): 1) \textit{\textbf{Rationale}}: the model's thinking process given the task. In this part, the model takes advantage of spatial and commonsense reasoning abilities in language-space to provide guidance for task completion. 2) \textit{\textbf{Action}}: Based on the provided rationale, the model generates appropriate coordinate-based actions. By aligning language and coordinates in the preceding stage, the rationale (articulated in language) can be effectively translated into coordinates (illustrated by the yellow to blue gradient dotted line in Figure \ref{fig:detailed-framework}-b) without the need for extensive fine-tuning data.

\begin{align}
    \label{equation-3} 
   & [\textbf{X}_v, \textbf{X}_\text{lang}] \xrightarrow[]{f_{\boldsymbol{\theta}} (\cdot)} 
  \overset{\raisebox{1.5ex}{\textit{\textbf{Action}}}}{\textbf{X}_\text{coor}} \quad \quad\quad\quad\quad\text{without rationale} \\
   \label{equation-4}
    & [\textbf{X}_v, \textbf{X}_\text{lang}] \xrightarrow[]{f_{\boldsymbol{\theta}}(\cdot)} [\overset{\raisebox{1.5ex}{\textit{\textbf{Rationale}}}}{\textbf{X}_\text{lang}},  \overset{\raisebox{1.5ex}{\textit{\textbf{Action}}}}{\textbf{X}_\text{coor}}] \quad\text{with rationale}
    \end{align}

Given this approach, a significant challenge lies in efficiently collecting high-quality rationale-action data pairs. This challenge arises from the need to generate data in the rationale-action sequence while maintaining consistency between them and ensuring the optimality of the generated actions. To address this, we designed a pipeline to automatically generate high-quality rationale-action data, as illustrated in Figure \ref{fig:rationale-collection}. Initially, given an image and a task instruction, a ground truth action is acquired from the simulator in a rule-based manner, ensuring the action's optimality. The ground truth action is then annotated on the input image, either as a trajectory or a subgoal point, depending on the task. Subsequently, a powerful vision-language model is employed to generate a rationale based on the action-labeled image and task instruction. It is crucial to note that there might be information leakage in the generated rationale, i.e., the information of the ground truth action directly appears in the generated rationale. To mitigate this, we introduce an additional constraint prompt, such as ``\textit{Your output should not include the ground truth action given in the image,}'' ensuring that the generated rationale remains valid.

\section{Experiments}

\subsection{Tasks}
\label{section:task_formulation}
Previous studies on spatial reasoning typically conduct open-loop evaluations on offline VQA datasets, which restricts the examination of their impact on downstream embodied tasks. To address these limitations, this paper adopts a more challenging setting by performing closed-loop evaluations within simulators. Additionally, we conduct offline evaluations of the fundamental capabilities of VLMs to investigate the relationship between these capabilities and end-to-end task planning abilities.

\paragraph{Closed-loop Embodied Task Planning}
\begin{table*}[t]
\small
\centering
\renewcommand{\arraystretch}{1.2}
\begin{tabular}{cc|cc|cc}
\Xhline{1.0pt}
\multicolumn{2}{c|}{\multirow{2}{*}{\textbf{Methods}}} & \multicolumn{2}{c|}{\textbf{Navigation Metrics}} & \multicolumn{2}{c}{\textbf{Manipulation Metrics}} \\ \cline{3-6} 
\multicolumn{2}{c|}{} & \multicolumn{1}{c|}{Distance Gain $\uparrow$} & Success Rate 
 $\uparrow$ & \multicolumn{1}{c|}{Collision Rate $\downarrow$} & Success Rate $\uparrow$ \\ \Xhline{1.0pt}
\multicolumn{2}{c|}{GPT-4o ICL} & \multicolumn{1}{c|}{-0.27} & 56.21 & \multicolumn{1}{c|}{65.20} & 0.00 \\ \hline
\multicolumn{2}{c|}{Llama3.2V 11B Zero-shot} & \multicolumn{1}{c|}{-2.47} & 54.73 & \multicolumn{1}{c|}{78.20} & 0.00 \\ \hline
\multicolumn{2}{c|}{RoboPoint 11B} & \multicolumn{1}{c|}{0.21} & 55.03 & \multicolumn{1}{c|}{88.80} &  0.00 \\ \Xhline{1.0pt}
\multicolumn{1}{c|}{\multirow{4}{*}{\begin{tabular}[c]{@{}c@{}}SpatialCoT\\(Ours)\end{tabular}}} & Direct Action Tuning & \multicolumn{1}{c|}{2.28} & 57.40 & \multicolumn{1}{c|}{21.35} & 75.81 \\ \cline{2-6} 
\multicolumn{1}{c|}{} & + Spatial Coordinate Alignment & \multicolumn{1}{c|}{3.23} & 60.65 & \multicolumn{1}{c|}{16.33} & 81.48 \\ \cline{2-6} 
\multicolumn{1}{c|}{} & + Chain-of-Thought Spatial Grounding & \multicolumn{1}{c|}{2.83} & 57.40 & \multicolumn{1}{c|}{18.51} & 77.78 \\ \cline{2-6} 
\multicolumn{1}{c|}{} & \begin{tabular}[c]{@{}c@{}}+ Spatial Coordinate Alignment\\ + Chain-of-Thought Spatial Grounding\end{tabular} & \multicolumn{1}{c|}{\textbf{3.33}} & \textbf{61.83} & \multicolumn{1}{c|}{\textbf{15.68}} & \textbf{82.57} \\ \Xhline{1.0pt}
\end{tabular}

\caption{\textbf{Results on closed-loop embodied task planning}: SpatialCoT demonstrates superior performance in both navigation and manipulation tasks, surpassing previous models, including both open-source and closed-source versions.}
\label{table:ablation-study-table}
\end{table*}

Inspired by the Goal-conditioned Markov Decision Process, we utilize this framework to break down the embodied task planning problem into varying levels of complexity: 
\begin{itemize}
    \item \textbf{State}: We consider occlusion as a primary factor, including visual, stacked, and encased occlusion. Other factors include object properties such as geometry and movability.
    \item \textbf{Goal}: This involves the number of objects, spatial constraints, and the abstraction level of the goal description.
    \item \textbf{Action}: The action space impacts the task's difficulty, including the format of actions and the number of required skills.
    \item \textbf{Transition}: This component addresses environmental transitions, encompassing uncertainty in dynamics.
\end{itemize}
In this work, we focus on a subset of the dimensions outlined above for simplicity. A comprehensive benchmark encompassing all dimensions will be provided in the future. This paper addresses two primary tasks: navigation and manipulation. For navigation tasks, unlike previous works, which treat navigation tasks as a region-localization problem (i.e., the model is asked to generate a position in the current view, such as ``\textit{Go to the position between the table and the chair}''), which do not require complex thinking and reasoning capabilities, this work adopts a more challenging setting. We use object-goal navigation as the evaluation task, where the agent must find an object not currently in view. The model is prompted to generate the best subgoal point in each observation to locate the target object as quickly as possible, such as ``Given the image, where would you go to find the \{target\_object\}?''. For manipulation tasks, we use tabletop rearrangement as the evaluation task, which is a more challenging extension of the Where2Place task in Robopoint. Given a target layout described in a language instruction, such as ``Set up a dining table for me,'' the model is tasked to move the objects step-by-step by generating the start and end positions for each object until the desired layout is achieved.

\paragraph{Fundamental Capabilities}
We also assess the fundamental capabilities of VLMs to understand their relationship with embodied task planning tasks. These capabilities are categorized, as detailed in Section~\ref{section:spatial-coordinate-alignment}, into four main categories: object understanding, affordance prediction, spatial relationships, and spatial compatibility. For further details, please refer to Section~\ref{section:spatial-coordinate-alignment}.

\subsection{Experimental Setup}
\paragraph{\textbf{Data Collection}} We collect data using two primary scene datasets. For navigation tasks, we utilize the Habitat Synthetic Scenes Dataset (HSSD) \cite{hssd} for data collection and employed Habitat~\cite{habitat3} as the simulator for closed-loop model evaluation. For manipulation tasks, we use Sapien \cite{sapien} as the simulator and generate diverse tabletop rearrangement tasks and data. With the power of large language models, the data construction process of tabletop rearrangement is semi-automated. Additionally, to improve visual fidelity and reduce the sim-to-real gap, we use Blender \cite{blender} as the renderer to obtain high-quality images for data collection. The amount of data and the examples for both stages are detailed in Appendix~\ref{appendix:dataset-examples}.

\paragraph{\textbf{Model Training}} For model training, we employed the Llama3.2-Vision 11B \cite{llama3.2V} as the backbone of the vision-language model. In both stages, we fine-tuned the model using LoRA \cite{lora} on the collected datasets. The training process spanned 2 epochs for each stage. Hyper-parameters used for model training can be found in Appendix~\ref{appendix:hyper-parameters}

\paragraph{\textbf{Baselines}}
We compare SpatialCoT with several baselines, including the specialized spatial reasoning model, RoboPoint, open-source VLMs such as LLaMA3.2V, and closed-source VLMs such as GPT-4o. The baselines include:
\begin{itemize}
    \item \textbf{RoboPoint}: While the vanilla RoboPoint is trained from a Vicuna-v1.5-13B \cite{vicuna} base model, for a fair comparison, we reproduce RoboPoint by fine-tuning a Llama3.2V-11B model (which shares the same backbone as our work) on the dataset provided in the original work.
    \item \textbf{LLama3.2V 11B Zeroshot}: This baseline evaluates the zero-shot spatial reasoning abilities of existing VLMs which are not specifically trained on spatial-related datasets.
    \item \textbf{Llama3.2V Direct-Action-Tuning}: To validate the effectiveness of the proposed method, we also introduce a basic baseline that is directly fine-tuned on action generation data.
    \item \textbf{GPT-4o}: We also compare our model with closed-source VLMs, using the current state-of-the-art model, OpenAI's GPT-4o, as the baseline.
\end{itemize}

\subsection{Experimental Results}

Our analysis aims to address the following questions: a) \emph{Is the two-stage training process effective in enhancing spatial reasoning ability of VLMs?} b) \emph{Which types of embodied planning tasks benefit most from improvements by SpatialCoT?} c) \emph{Is there a positive correlation between the fundamental capabilities of VLMs and their downstream performance in embodied planning tasks?} d) \emph{How does chain-of-thought contribute to improving the spatial reasoning capabilities of VLMs?}

\paragraph{\textbf{Question 1: Is the two-stage training process effective in enhancing spatial reasoning ability of VLMs?}}
We compare SpatialCoT with baseline models across navigation and manipulation tasks, as described in the preceding sections. Additionally, we conduct ablation experiments to verify the effectiveness of the two-stage approach.

\textbf{Results on Navigation Tasks} For navigation tasks, we introduce two metrics as following:
\begin{itemize}
    \item \textbf{Distance Gain (DG)}: This metric evaluates the quality of generated actions at each step by calculating the negative traverse distance from the generated position to the nearest target object, using a path planning algorithm. We also perform mean normalization across all possible actions to highlight the relative improvements achieved. Let $a$ represent the generated action, $a_i$ one of the possible actions, $g$ the position of the target object, and $d(\cdot)$ the traverse distance function. The formula is: $DG= -d(a, g) + \sum_{i=1}^{N}d(a_i, g) / N)$.  When the generated position is closer to the target object, the score is higher, and when the generated position is farther away, the score is lower.
    \item \textbf{Success Rate (SR)}: This metric provides a closed-loop evaluation of the model's performance within a simulator. At each timestep, given the current observation, the model has the option to select a subgoal within the present view or adjust the view to explore the environment further if needed. Subsequently, a low-level controller executes the generated actions and gathers the next observations for decision-making. This iterative process continues until the model reaches the maximum number of steps, set at 500 in this study.
\end{itemize}
\begin{figure*}
    \centering
    \includegraphics[width=\textwidth]{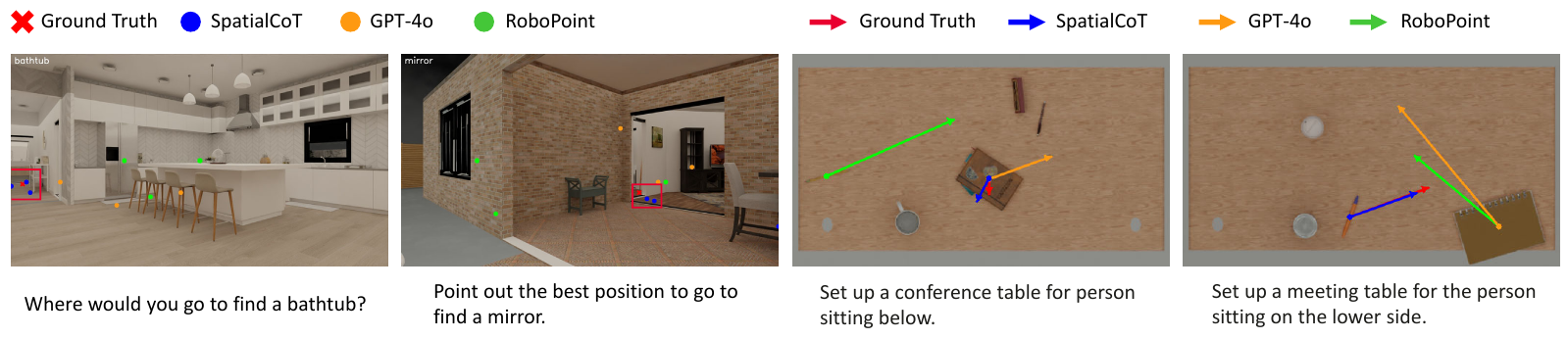}
    \vspace{-10pt} 
    \caption{Visualization of spatial reasoning results on navigation and manipulation tasks.}
    \label{fig:case_show}
\end{figure*}

\begin{figure*}
    \centering
    \includegraphics[width=1.0\textwidth]{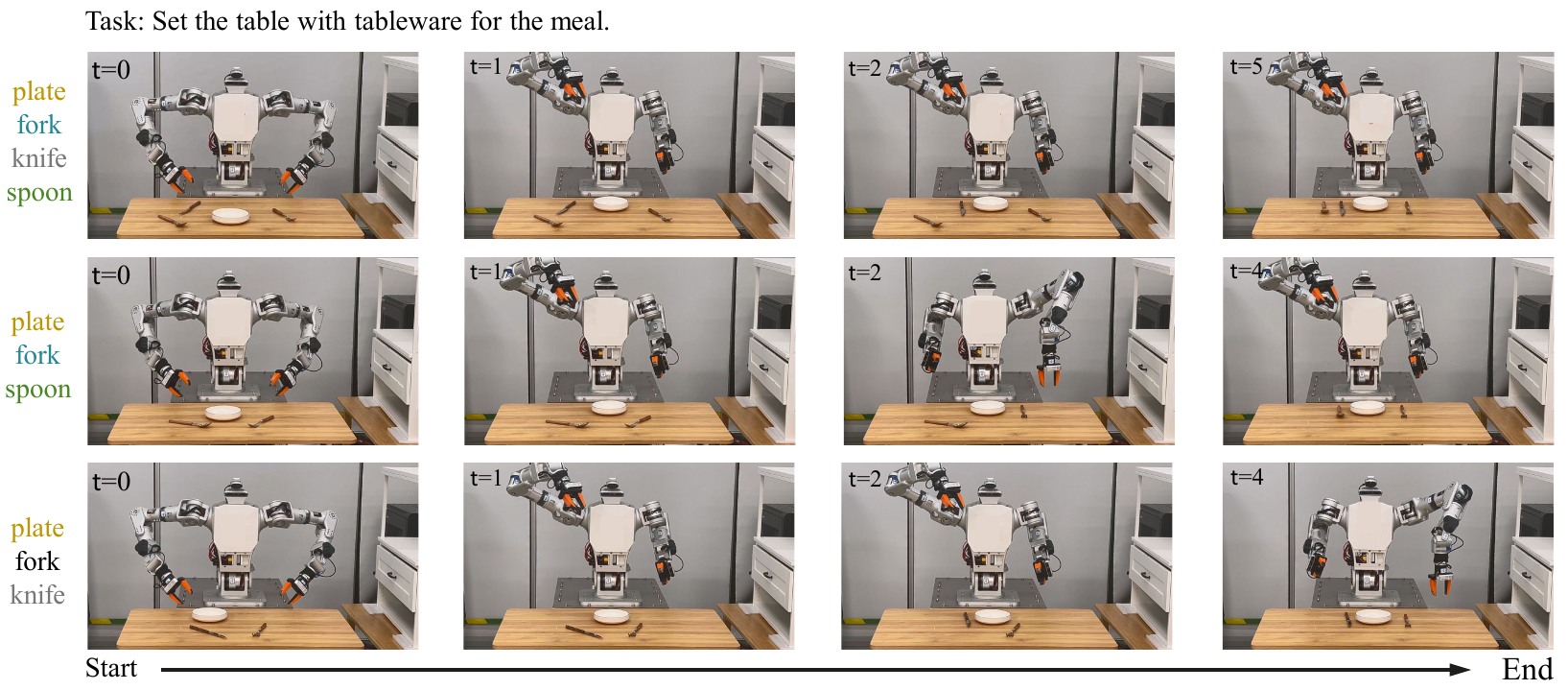}
    \vspace{-10pt} 
    \caption{\textbf{Real-world rearrangement experiments}: SpatialCoT arranges various object combinations into reasonable layouts, adhering to physical constraints and avoiding collisions.}
    \label{fig:realworld-experiments}
\end{figure*}

As illustrated in Table~\ref{table:ablation-study-table}, the distance gains for GPT-4o ICL and LLama3.2V 11B Zero-shot are -0.27 and -2.47, respectively, indicating that the quality of generated actions are below average. RoboPoint achieves a distance gain (DG) of 0.21, demonstrating that VLMs trained on typical spatial reasoning tasks are insufficient for addressing more complex tasks requiring higher reasoning abilities. We also implement a baseline approach involving fine-tuning the model directly on the action generation data, resulting in a DG of 2.28. With the \emph{spatial coordinate bi-directional alignment}, the DG improves to 3.23, and with \emph{chain-of-thought spatial grounding}, it improves to 2.83. Combining both stages results in a DG of 3.33, indicating a significant improvement over direct action tuning (+ 46\% relative improvement). Regarding the success rate, SpatialCoT achieves 61.83\%, which is an increase of 4.43\% compared to direct action tuning, and also the highest success rate among all evaluated open-source and closed-source models.

\textbf{Results on Manipulation Tasks}: For tabletop rearrangement, we introduce two metrics:
\begin{itemize}
    \item \textbf{Collision Rate (CR)}: This metric assesses the validity of the generated action. An object cannot collide with other objects during the task. The task fails if a collision occurs. 
    \item \textbf{Success Rate (SR)}: The task is considered successful if the layout of the objects meets the requirements described in the instructions and no collision occurs during the process; otherwise, the task is deemed a failure.
\end{itemize}
As illustrated in Table \ref{table:ablation-study-table}, previous models did not succeed in zero-shot evaluation. This failure is primarily due to the complexity of our tabletop rearrangement tasks, which require an understanding of physical concepts (such as collisions) and human common sense (such as when to stop). Additionally, these tasks demand long-horizon planning capabilities. Consequently, models that use zero-shot evaluation struggle to complete overall task planning effectively. However, direct action tuning significantly reduces the collision rate to 21.3\% while achieving a success rate of 75.8\%. SpatialCoT further enhances these metrics, achieving a collision rate of 15.6\% and a success rate of 82.6\%. This demonstrates a notable improvement in the end-to-end success rate (an increase of 6.8\%) and a reduction in the collision rate (a decrease of 5.6\%).

We visualize the results of SpatialCoT and the baselines, as shown in Figure~\ref{fig:case_show}. Beyond simulation, we evaluate our model in real-world scenarios using a dual-arm robot. As demonstrated in Figure \ref{fig:realworld-experiments}, our model exhibits impressive transferability.

\paragraph{\textbf{Question 2: Which types of embodied planning tasks benefit most from improvements by SpatialCoT?}} We analyzed the results by categorizing the tasks into several levels, based on the principle described in Section~\ref{section:task_formulation}. The results are presented in Table~\ref{table:task_levels_manipulation} and Table~\ref{table:task_levels_navigation}. 
Table~\ref{table:task_levels_manipulation} reveals that the majority of failures in manipulation tasks stem from non-unique objects (level-3) and a high number of objects (level-4). This results in crowded scenes and an increased likelihood of collisions. Our model demonstrated significant improvements in these tasks (+8.68\% and +20.00\% respectively), indicating an enhanced understanding of object relationships and physical cognitive abilities, such as collision avoidance.
In navigation tasks, SpatialCoT exhibited superior performance at levels 1, 2, and 4. The most notable improvement was at the most challenging level (+8.66\%), level 4, characterized by fewer goals and greater distances. These results suggest that SpatialCoT effectively enhances the model's capability to manage tasks with sparse reward signals, requiring advanced spatial understanding and reasoning.

\begin{table}[]
\scriptsize
\renewcommand{\arraystretch}{1.3}
\centering

\begin{tabular}{c|c|c|c|cc}
\hline
\multirow{2}{*}{\begin{tabular}[c]{@{}c@{}}Manipulation\\ Task Levels\end{tabular}} & \multirow{2}{*}{\begin{tabular}[c]{@{}c@{}}Unique\\ Objects\end{tabular}} & \multirow{2}{*}{\begin{tabular}[c]{@{}c@{}}Stacked\\ Objects\end{tabular}} & \multirow{2}{*}{\begin{tabular}[c]{@{}l@{}}Objects\\ Number\end{tabular}} & \multicolumn{1}{c|}{\ DAT \ } & SpatialCoT \\ \cline{5-6} 
 &  &  &  & \multicolumn{2}{c}{Success Rate} \\ \hline
Level 1 & \checkmark & × & $\leq3$ & \multicolumn{1}{c|}{95.00} & \textbf{95.00} \\ \hline
Level 2 & \checkmark & \checkmark & 4$\sim$5 & \multicolumn{1}{c|}{86.11} & \textbf{90.56} \\ \hline
Level 3 & × & \checkmark & 6$\sim$8 & \multicolumn{1}{c|}{68.49} & \textbf{77.17} \\ \hline
Level 4 & × & \checkmark & $\geq9$ & \multicolumn{1}{c|}{25.00} & \textbf{45.00} \\ \hline
\end{tabular}

\caption{Results on manipulation tasks across different difficulty levels, with DAT representing direct action tuning.}
\label{table:task_levels_manipulation}
\end{table}

\begin{table}[]
\scriptsize
\renewcommand{\arraystretch}{1.3}
\centering
\begin{tabular}{c|c|c|cc}
\hline
\multirow{2}{*}{\begin{tabular}[c]{@{}c@{}}Navigation\\ Task Levels\end{tabular}} & \multirow{2}{*}{\begin{tabular}[c]{@{}c@{}}Number\\ of Goals\end{tabular}} & \multirow{2}{*}{\begin{tabular}[c]{@{}c@{}}Distance\\ to Goal\end{tabular}} & \multicolumn{1}{c|}{\ DAT \ } & SpatialCoT \\ \cline{4-5} 
 &  &  & \multicolumn{2}{c}{Success Rate} \\ \hline
Level 1 & $>2$ & $\leq4.5m$ & \multicolumn{1}{c|}{80.77} & \textbf{82.69} \\ \hline
Level 2 & $\leq2$ & $\leq4.5m$ & \multicolumn{1}{c|}{83.33} & \textbf{86.11} \\ \hline
Level 3 & $>2$ & $>4.5m$ & \multicolumn{1}{c|}{\textbf{71.26}} & 70.11 \\ \hline
Level 4 & $\leq2$ & $>4.5m$ & \multicolumn{1}{c|}{38.67} & \textbf{47.33} \\ \hline
\end{tabular}

\caption{Results on navigation tasks across different difficulty levels, with DAT representing direct action tuning.}
\label{table:task_levels_navigation}

\end{table}


\paragraph{\textbf{Question 3: Is there a positive correlation between the fundamental capabilities of VLMs and their downstream performance in embodied planning tasks?}}

\begin{figure}[ht]
    \centering
    \begin{subfigure}[b]{0.48\linewidth}
        \centering
        \includegraphics[width=0.9\linewidth]{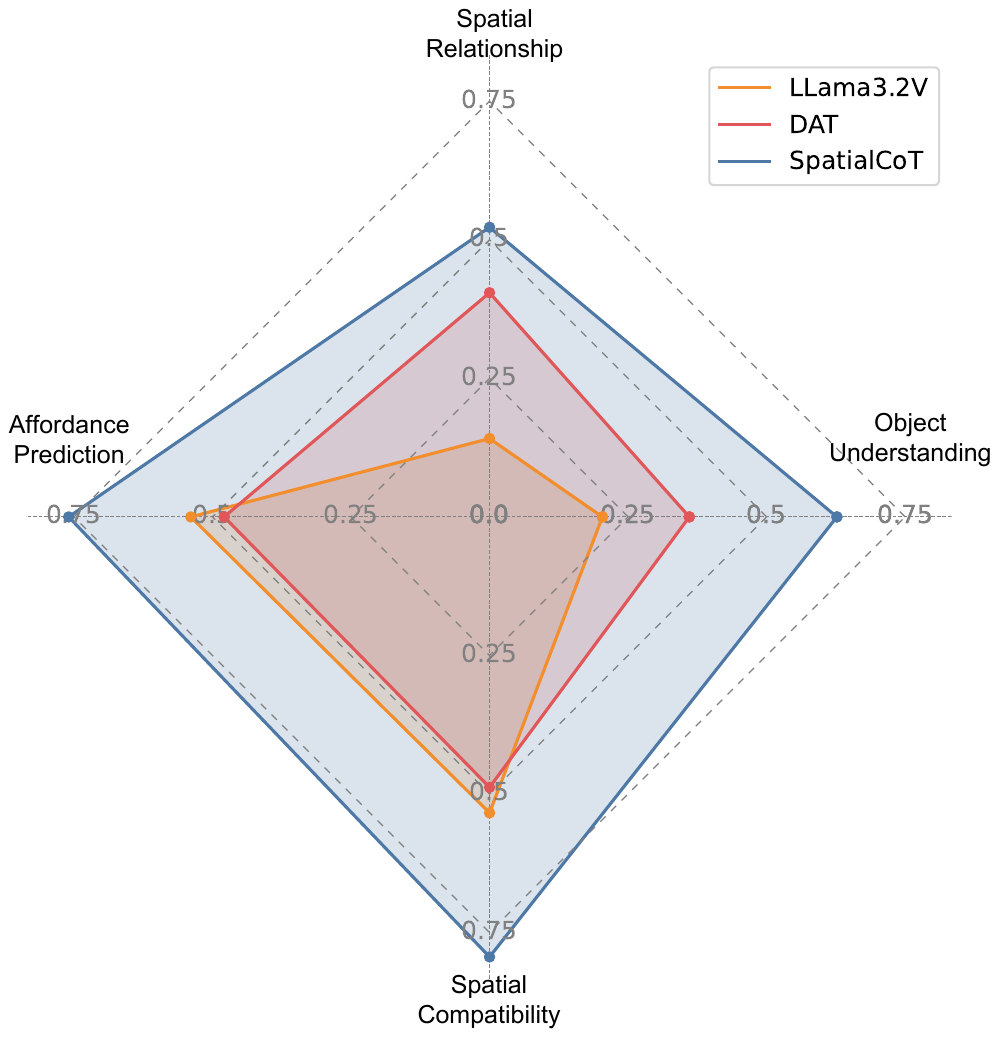}
        \caption{Fundamental capabilities}
        \label{fig:correlation-sub1}
    \end{subfigure}
    \hspace{0.01\linewidth}
    \begin{subfigure}[b]{0.48\linewidth}
        \centering
        \includegraphics[width=0.95\linewidth]{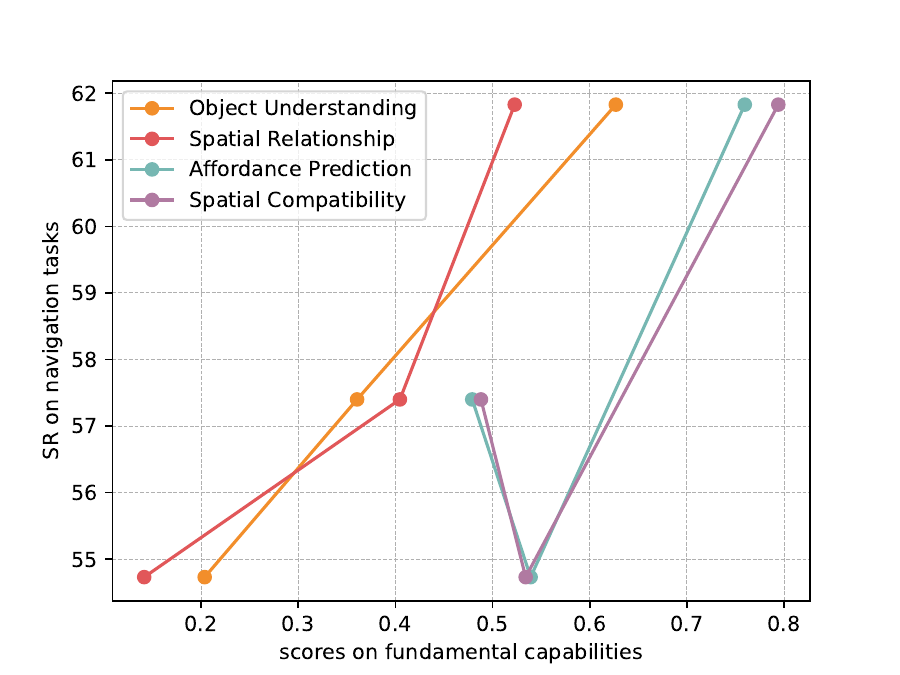}
        \caption{Correlation chart}
        \label{fig:correlation-sub2}
    \end{subfigure}
    \caption{Analysis of the correlation between fundamental capabilities and embodied task planning of VLMs, with DAT representing direct action tuning.}
    \label{fig:correlation}
\end{figure}

In our evaluation of the fundamental capabilities of Vision-Language Models (VLMs), we find that SpatialCoT consistently outperforms other models across all evaluated categories (see Figure~\ref{fig:correlation-sub1}). To further explore the correlation between each category of fundamental capability and downstream performance, we present these correlations in Figure~\ref{fig:correlation-sub2}. The horizontal axis represents the scores of fundamental capabilities, while the vertical axis shows the success rates in embodied planning tasks. Each line corresponds to a specific category of fundamental capability. The results reveal a clear positive relationship between object understanding and spatial relationships (orange and red lines). The other two categories also display a positive correlation trend, though not entirely monotonic. These findings demonstrate that there is a positive correlation between the fundamental capabilities of VLMs and their downstream performance, providing a valuable basis for further research in this field.

\begin{figure}
    \centering
    \includegraphics[width=0.9\linewidth]{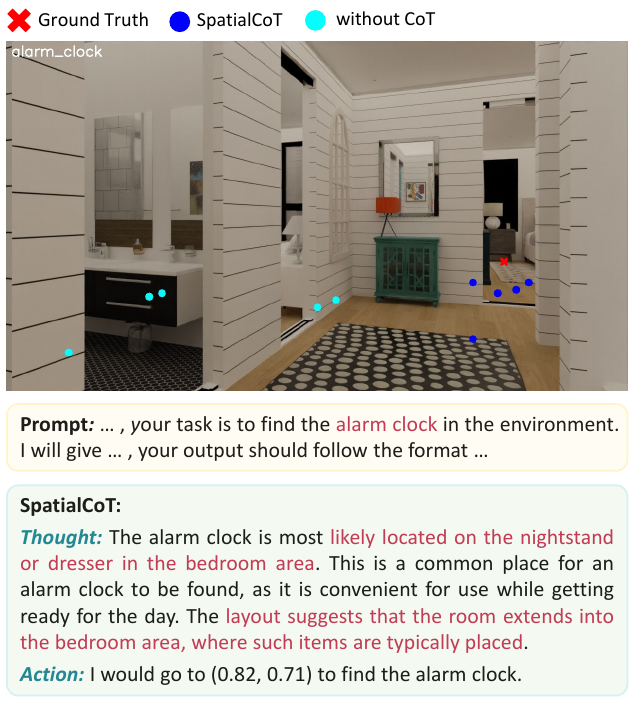}
    \caption{Case study for chain-of-thought spatial grounding}
    \label{fig:cot-case-study}
\end{figure}

\paragraph{\textbf{Question 4: How does chain-of-thought contribute to improving the spatial reasoning capabilities of VLMs?}}
Experimental results demonstrate that the chain-of-thought process significantly enhances the model's ability to utilize spatial and contextual information, such as room layout and commonsense knowledge, to arrive at the correct answer. To illustrate this, we present a case study (see Figure~\ref{fig:cot-case-study}). In this task, the model is instructed to locate an alarm clock within a house. The SpatialCoT model first considers the typical location of an alarm clock, infers the bedroom's position based on the current layout, and ultimately produces accurate results. In contrast, the baseline model (without CoT) generates disordered results throughout the room.
\\

\section{Limitations}

SpatialCoT employs coordinate-based actions for embodied task planning; however, it does not account for complex actions such as rotations, rendering it unable to manage tasks that require object rotation. Moreover, as a vision-language model, SpatialCoT relies on 2D images for visual input, thereby necessitating future research to explore the potential of 3D inputs, especially in large spaces.
\section{Conclusion}

In this paper, we propose a novel approach, SpatialCoT, designed to enhance the spatial reasoning capabilities of vision-language models through a two-stage training paradigm: spatial coordinate bi-directional alignment and chain-of-thought spatial grounding. By explicitly leveraging the language-based reasoning abilities of vision-language models and anchoring them in coordinate-based actions, our approach significantly improves the model's performance in handling complex embodied tasks. Experimental results demonstrate that SpatialCoT outperforms previous methods in challenging embodied tasks, including navigation and manipulation.



\clearpage
\newpage

\section{Appendix}  

\definecolor{darkgreen}{rgb}{0.0, 0.5, 0.0}
\definecolor{blueviolet}{rgb}{0.54, 0.17, 0.89}

\subsection{Dataset Examples}
\label{appendix:dataset-examples}

\subsubsection{Dataset for Spatial Coordinate Bi-directional Alignment}

Here we show some examples of the data for spatial coordinate alignment.

\begin{tcolorbox}[colback=blue!1!white, colframe=blue!5!white, title=\textcolor{black}{\textbf{Object Understanding}}, breakable]
\begin{center}
    \includegraphics[width=1.0\textwidth]{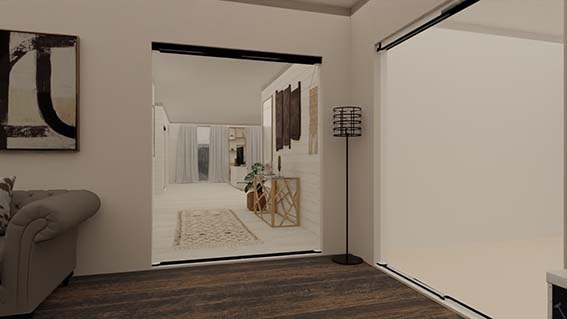}
\end{center}
\rule{\linewidth}{1.0pt} 
\textbf{\textcolor{violet}{Coordinates Understanding}} \\
\textbf{Prompt}: \textit{Given a location on the image, you should recognize the object on this location. The location will be formatted as (x, y), with each coordinate ranging from 0 to 1 and rounded to two decimal places. Direct output the object name, no additional reasoning process is needed. \textcolor{violet}{Recognize the object located at (0.06, 0.78)}.} \\
\textbf{Response}: \textit{couch} \\
\rule{\linewidth}{0.8pt} 
\textbf{\textcolor{darkgreen}{Coordinates Generation}} \\
\textbf{Prompt}: \textit{\textcolor{darkgreen}{Identify all instances of couch in the provided image}. For each detected instance, provide the center location of the object. The predicted location should be formatted as (x, y), with each coordinate ranging from 0 to 1 and rounded to two decimal places. Your output should be in the following format: Detected couch(s): [(x1, y1), (x2, y2), ...]} \\
\textbf{Response}: \textit{Detected couch(s): [(0.06, 0.78)]} 
\end{tcolorbox}

\begin{tcolorbox}[colback=darkgreen!1!white, colframe=darkgreen!5!white, title=\textcolor{black}{\textbf{Affordance Prediction}}, breakable]
\begin{center}
    \includegraphics[width=1.0\textwidth]{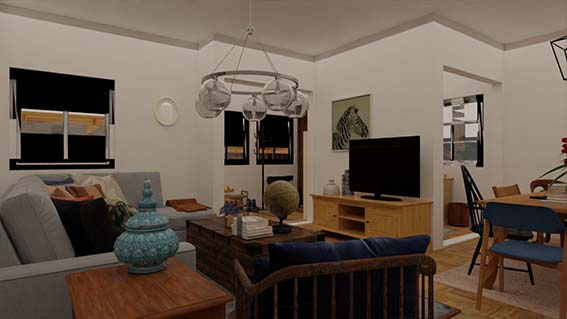}
\end{center}
\rule{\linewidth}{1.0pt} 
\textbf{\textcolor{violet}{Coordinates Understanding}} \\
\textbf{Prompt}: \textit{Given a location on the image, determine if it is navigable for the robot (i.e., on the ground). The location will be formatted as (x, y), with each coordinate ranging from 0 to 1 and rounded to two decimal places. Provide a direct answer with ``yes'' or ``no'' without additional reasoning. \textcolor{violet}{Is location (0.45, 0.97) navigable?}} \\
\textbf{Response}: \textit{no} \\
\rule{\linewidth}{0.8pt} 
\textbf{\textcolor{darkgreen}{Coordinates Generation}} \\
\textbf{Prompt}: \textit{\textcolor{darkgreen}{Given an image, generate a navigable point for the robot}. The output should will be formatted as (x, y), with each coordinate ranging from 0 to 1 and rounded to two decimal places. Please respond directly with the generated point. No additional reasoning is required.} \\
\textbf{Response}: \textit{(0.80, 0.85)} 
\end{tcolorbox}

\begin{tcolorbox}[colback=brown!3!white, colframe=brown!20!white, title=\textcolor{black}{\textbf{Spatial Relationship}}, breakable]
\begin{center}
    \includegraphics[width=1.0\textwidth]{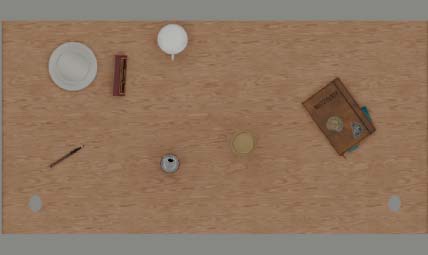}
\end{center}
\rule{\linewidth}{1.0pt} 
\textbf{\textcolor{violet}{Coordinates Understanding}} \\
\textbf{Prompt}: \textit{\textcolor{violet}{Describe the spatial relationship between the objects located at coordinates (0.17, 0.26) and (0.40, 0.15)}, with each coordinate ranging from 0 to 1 and rounded to two decimal places.} \\
\textbf{Response}: \textit{The plate is positioned to the lower left of the mug.} \\
\rule{\linewidth}{0.8pt} 
\textbf{\textcolor{darkgreen}{Coordinates Generation}} \\
\textbf{Prompt}: \textit{Given the image, point out the object located to the left
of the notebook. The output should be formatted as (x, y), with each coordinate ranging from 0 to 1 and rounded to two decimal places. Please respond directly with the generated point. No additional reasoning is required.} \\
\textbf{Response}: \textit{(0.57, 0.55)} 
\end{tcolorbox}

\begin{tcolorbox}[colback=violet!3!white, colframe=violet!20!white, title=\textcolor{black}{\textbf{Spatial Compatibility}}, breakable]
\begin{center}
    \includegraphics[width=1.0\textwidth]{images/table1.jpg}
\end{center}
\rule{\linewidth}{1.0pt} 
\textbf{\textcolor{violet}{Coordinates Understanding}} \\
\textbf{Prompt}: \textit{Given the image, determine whether the object will collide with other objects after moving it from the initial position to the target position. The point will be formatted as (x, y), with each coordinate ranging from 0 to 1 and rounded to three decimal places. Provide a direct answer with 'yes' or 'no' without additional reasoning. \textcolor{darkgreen}{Will a collision occur after moving the plate from (0.17, 0.26) to (0.70, 0.52)?}} \\
\textbf{Response}: \textit{yes} \\
\rule{\linewidth}{0.8pt} 
\textbf{\textcolor{darkgreen}{Coordinates Generation}} \\
\textbf{Prompt}: \textit{\textcolor{darkgreen}{Generate a collision-free location for the notebook}. The output should be formatted as (x, y), with each coordinate ranging from 0 to 1 and rounded to two decimal places. Please respond directly with the generated point. No additional reasoning is required.} \\
\textbf{Response}: \textit{(0.57, 0.55)
} 
\end{tcolorbox}

\begin{table}[h]
\small
\renewcommand{\arraystretch}{1.3}
\centering
\begin{tabular}{c|c|c|c}
\hline
Data Type & \begin{tabular}[c]{@{}c@{}}Coordinates\\ Understanding\end{tabular} & \begin{tabular}[c]{@{}c@{}}Coordinates\\ Generation\end{tabular} & Total \\ \hline
Object Understanding & 137k & 127k & 264k \\ \hline
Affordance Prediction & 40k & 24k & 64k \\ \hline
Spatial Relationship & 40k & 40k & 80k \\ \hline
Spatial Compatibility & 40k & 40k & 80k \\ \hline
Total & 257 & 231 & 488k \\ \hline
\end{tabular}
\caption{Data for spatial coordinate bi-directional alignment}
\label{table:stage1_data1}
\end{table}

\definecolor{darkblue}{rgb}{0.0, 0.0, 0.55}

\subsubsection{Dataset for Chain-of-Thought Grounding}

Here we show some examples of the data for chain-of-thought grounding.

\begin{tcolorbox}[colback=gray!3!white, colframe=gray!20!white, title=\textcolor{black}{\textbf{Navigation Tasks}}, breakable]
\begin{center}
    \includegraphics[width=1.0\textwidth]{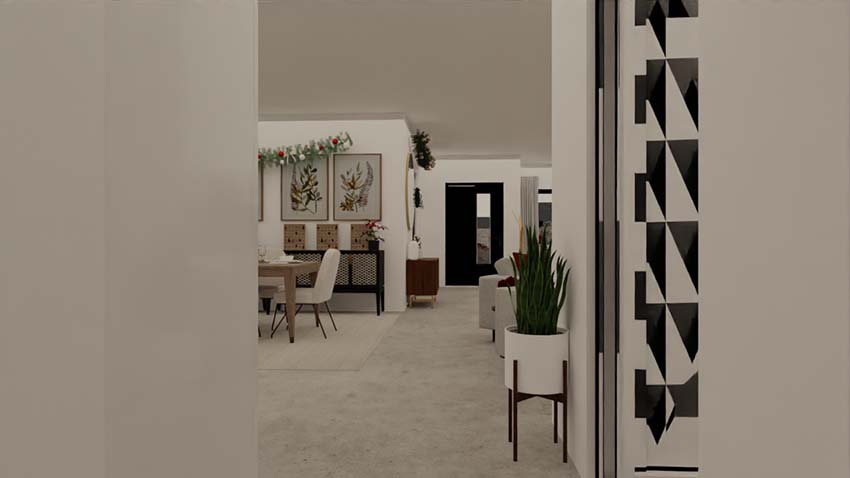}
\end{center}
\rule{\linewidth}{1.0pt} 
\textbf{Without Rationale} \\
\textbf{Prompt}: \textit{You are a robot in an unfamiliar environment. Your task is to find the dishwasher in the environment. Based on the image, predict the optimal location to move next to find the dishwasher. The predicted location should be in the format (x, y), with each number ranging from 0 to 1 and rounded to two decimal places. Ensure the predicted location is navigable (i.e., on the ground). Please respond directly with: ``I should go to (x, y) to find the dishwasher.'' No additional reasoning is required.} \\
\textbf{Response}: \textit{I should go to (0.30, 0.82) to find the dishwasher.} \\
\rule{\linewidth}{0.8pt} 
\textbf{\textcolor{blue}{With Rationale}} \\
\textbf{Prompt}: \textit{You are a robot in an unfamiliar environment. Your task is to find the dishwasher in the environment. I will give you a current observation image. Based on this image, predict the optimal next move to find the dishwasher. The predicted location should be formatted as (x, y), with each coordinate ranging from 0 to 1 and rounded to two decimal places. Ensure the predicted location is navigable (i.e., on the ground). Output your thinking process before getting the final answer. Your output should in the following format: \\ \textcolor{blue}{Thought}: [Put your thinking process there. You should think about the location of the target object or the region where it is located. This can be achieved by reasonably imagining the unseen areas based on the room layout] \\ \textcolor{blue}{Action}: [I should go to (x, y) to find the dishwasher.]} \\
\textbf{Response}: \\ \textit{\textcolor{blue}{Thought}: The dishwasher is most likely located in the kitchen area. Given the layout, the kitchen could be adjacent to the dining area. Following the path, it is reasonable to assume that the kitchen is to the left of the dining area, possibly behind the wall on the left side of the image. This is a common layout in many homes where the kitchen is near the dining area for convenience. \\ \textcolor{blue}{Action}: I should go to (0.30, 0.82) to find the dishwasher.
} 
\end{tcolorbox}

\begin{table}[h]
\small
\renewcommand{\arraystretch}{1.3}
\centering
\begin{tabular}{c|c|c|c}
\hline
Data Type & without Rationale & with Rationale & Total \\ \hline
Navigation Tasks & 50k & 15k & 65k \\ \hline
Manipulation Tasks & 260k & 260k & 520k \\ \hline
Total & 310k & 275k & 585k \\ \hline
\end{tabular}
\caption{Data for chain-of-thought spatial grounding}
\label{table:stage2_data}
\end{table}

\subsection{Hyper-parameters for Model Training}
\label{appendix:hyper-parameters}
\begin{table}[H]
\small
\renewcommand{\arraystretch}{1.3}
\centering
\begin{tabular}{c|c}
\hline
\textbf{Hyper-parameters} & \textbf{Value} \\ \hline
learning\_rate & 1e-5 \\ \hline
learning\_rate\_decay (per epoch) & 0.9 \\ \hline
weight\_decay & 0 \\ \hline
gradient\_clipping & False \\ \hline
batch\_size & $16\times8$ \\ \hline
batching\_strategy & padding \\ \hline
gradient\_accumulation\_steps & 1 \\ \hline
num\_epochs & 2 \\ \hline
peft\_method & lora \\ \hline
freeze\_layers & False \\ \hline
enable\_fsdp & True \\ \hline
use\_fp16 & False \\ \hline
mixed\_precision & True \\ \hline
lora\_alpha & 32 \\ \hline
lora\_dropout & 0.05 \\ \hline
lora\_r & 8 \\ \hline
\end{tabular}
\caption{Hyper-parameters for Model Training}
\end{table}


\end{document}